\documentclass[letterpaper, 10 pt, conference]{ieeeconf}  % Comment this line out if you need a4paper
\IEEEoverridecommandlockouts                              % This command is only needed if 
\usepackage{graphicx}
\usepackage{caption}
\usepackage[english]{babel}
\usepackage{amsmath,amssymb,enumerate}
\usepackage{amsfonts}
\usepackage{mathtools}
\usepackage[mathscr]{euscript}
\usepackage{mathrsfs}
\usepackage{times}
\usepackage[usenames,dvipsnames,svgnames,table, xcdraw]{xcolor} %http://ctan.org/pkg/xcolor
\usepackage{url}
\usepackage{subcaption}
\usepackage{booktabs}
\usepackage{multirow}
\usepackage[ruled,vlined,linesnumbered]{algorithm2e}
\SetKw{Continue}{continue}
\SetKw{Break}{break}
\let\labelindent\relax
\usepackage{enumitem}
\usepackage{cite} % to group references
\usepackage[utf8]{inputenc}
\usepackage{balance}
\usepackage[normalem]{ulem}

\newif\ifshowcomment
\newif\ifnumberrevision
\newif\ifcolorrevision
\newif\ifstrikeremovel

\showcommenttrue
\numberrevisiontrue
\colorrevisiontrue
\strikeremoveltrue

\newcommand{\Acal}{\mathcal{A}}
\newcommand{\Bcal}{\mathcal{B}}

\newcommand{\transpose}{\mathsf{T}}

\newcommand{\node}{N}

\newcommand{\tableref}{Table}

\newcommand{\numC}{|C|}
\newcommand{\numK}{|K|}
\newcommand{\numM}{|M|}
\newcommand{\numV}{n_k}

\usepackage[breaklinks=true,letterpaper=true,colorlinks,bookmarks=false]{hyperref}

\title{\LARGE \bf Learning Task Requirements and Agent Capabilities for Multi-agent Task Allocation}

\author{Bo Fu$^{1}$, William Smith$^{2}$, Denise Rizzo$^{2}$, Matthew Castanier$^{2}$, Maani Ghaffari$^{1}$, and Kira Barton$^{1}$%
\thanks{This work was supported by the Automotive Research Center, a U.S. Army center of excellence for modeling and simulation of ground vehicles. Distribution Statement A: Approved for public release; distribution unlimited (OPSEC 6859).}
\thanks{$^{1}$Bo Fu, Maani Ghaffari, and Kira Barton are with the Department of Robotics, University of Michigan, Ann Arbor, MI 48109, USA {\tt\small \{bofu, maanigj, bartonkl\}@umich.edu } }
\thanks{$^{2}$William Smith, Denise Rizzo, and Matthew Castanier are with the US Army DEVCOM Ground Vehicle Systems Center, Warren, MI 48397, USA  {\tt\small \{william.c.smith1019, denise.m.rizzo2, matthew.p.castanier\}.civ@army.mil}
}%
}

\begin{document}

% Override seps
\setlength{\textfloatsep}{\baselineskip}
\setlength{\floatsep}{\baselineskip}
\setlength{\intextsep}{\baselineskip}
\setlength{\dbltextfloatsep}{\baselineskip}
\setlength{\dblfloatsep}{\baselineskip}

% \floatsep 1\baselineskip plus  0.2\baselineskip minus  0.2\baselineskip
% \intextsep 1\baselineskip plus 0.2\baselineskip minus  0.2\baselineskip

\maketitle
\thispagestyle{empty}
\pagestyle{empty}
\begin{abstract}

This paper presents a learning framework to estimate an agent capability and task requirement model for multi-agent task allocation.
With a set of team configurations and the corresponding task performances as the training data, linear task constraints can be learned to be embedded in many existing optimization-based task allocation frameworks.
Comprehensive computational evaluations are conducted to test the scalability and prediction accuracy of the learning framework with a limited number of team configurations and performance pairs.
A ROS and Gazebo-based simulation environment is developed to validate the proposed requirements learning and task allocation framework in practical multi-agent exploration and manipulation tasks.
Results show that the learning process for scenarios with 40 tasks and 6 types of agents uses around 12 seconds, ending up with prediction errors in the range of 0.5-2\%.

\end{abstract}

% \begin{IEEEkeywords}
% Some keywords
% \end{IEEEkeywords}

\section{Introduction}
With recent advances in robotic perception, motion planning, and control, robots have become capable of conducting a series of tasks including autonomous driving \cite{ding2021epsilon}, exploration \cite{cai2021non, fu2022simultaneous},
% surveillance \cite{yu2019coverage, sung2020distributed},
and manipulation \cite{werfel2014designing, chou2020explaining}. This energizes the field of multi-agent systems where robots are organized in teams to conduct larger and more complex tasks \cite{korsah2013comprehensive, rizk2019cooperative}.

Apart from the planning and control of single-robot motions, fundamental problems in a multi-agent system include resource allocation, task assignment, and scheduling, i.e. identifying how many agents are needed for a task, determining which agent individual should perform a specific task requirement, and deriving a plan that the agents should follow to complete the tasks. In a multi-agent system, these three problems are considered in an optimization-based or a reinforcement learning-style framework \cite{banks2020multi, aksaray2015distributed, fu2020heterogeneous, fu2021robust, ravichandar2020strata, prorok2017impact, messing2022grstaps, liemhetcharat2011modeling, liemhetcharat2012weighted}.

\begin{figure*}[t]
    \centering
    \includegraphics[width=\linewidth]{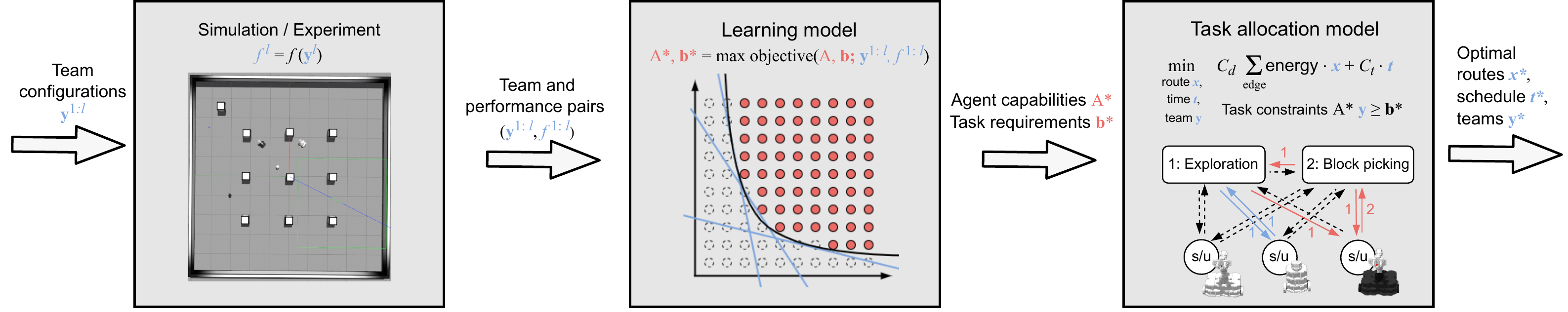}
    \caption{System architecture. The architecture comprises: a simulation/experimental environment (Sec. \ref{sec:result_simulation}), a learning model that inputs team configurations and performances and learns agent capabilities and task requirements (Sec. \ref{sec:learning_model}), and a task allocation model that generates an optimal teaming plan based on the learned capabilities and task requirements (Sec. \ref{sec:task_allocation_model}).
    In the graph for the task allocation model, there are two task nodes, and s/u means the start and terminal nodes for the three agent types. The edges are highlighted if there is a number (also highlighted) of agents traveling the edge.
    }
    \label{fig:main_diagram}
\end{figure*}

While optimization model-based systems are usually scalable, generalizable, and require no training, they often assume the modeling parameters are known or can be provided \cite{fu2020heterogeneous, fu2021robust, ravichandar2020strata, prorok2017impact, messing2022grstaps, liemhetcharat2011modeling, liemhetcharat2012weighted}. However, in practice, model parameters are often unknown or contain significant uncertainty. Therefore, most existing optimization-based approaches are limited by this assumption. 
This work presents a method to learn the parameters of a task requirement and agent capability model, which can be applied in a task allocation problem to represent the required resources of each task and easily embedded as constraints in the optimization \cite{fu2020heterogeneous, fu2021robust, ravichandar2020strata, prorok2017impact}.
The procedure can be shown to be applied to different types of tasks (generalizable) when the assumption of a cumulative capability and known sparsity pattern, provided in Sec. \ref{sec:learning_model}, is satisfied.
The system architecture is summarized in Fig. \ref{fig:main_diagram} and will be discussed in the following sections. This paper provides the following contributions.

\begin{enumerate}[label={\arabic*)}]
    \item The development of a generalizable framework with training data gathering, task requirement learning, and task allocation planning for multi-agent tasks.
    \item The formulation of a linear program that learns the parameters of a task requirement and agent capability model for task allocation.
    \item A comprehensive computational investigation that evaluates the scalability, accuracy, and speed of the proposed learning program.
    \item The implementation of a ROS and Gazebo-based multi-agent simulation that validates the proposed framework through a practical case study. Open-source code:\\ {\footnotesize
    \urlstyle{same}
    % \url{https://gitlab.com/barton-research-group/open/learn-multiagent-taskreq}
    \url{https://brg.engin.umich.edu/publications/learn-multiagent-taskreq/}
    % \href{https://gitlab.com/barton-research-group/open/learn_multiagent_taskreq}{https://gitlab.com/barton-research-group/open/learn\_multiagent\_taskreq}
    }
\end{enumerate}

\section{Related Work}\label{sec:related_work}

\subsection{Optimization model-based Multi-agent Task Allocation}\label{sec:model_based_task_allocation}

Multi-agent task allocation is the problem of determining which agents should execute a given task to achieve the overall system goals \cite{gerkey2004formal, korsah2013comprehensive}.
% The task allocation is further defined by the requirements on agent coordination, heterogeneity of agents and tasks, and dependency between different tasks.

Both learning-based \cite{elfakharany2020towards, baker2019emergent, li2020graph, wang2020mobile, wu2021impact} and optimization model-based frameworks \cite{banks2020multi, aksaray2015distributed, fu2020heterogeneous, fu2021robust, ravichandar2020strata, prorok2017impact, messing2022grstaps, liemhetcharat2011modeling, liemhetcharat2012weighted} have been applied to the task allocation problem. Learning-based frameworks usually require low computational costs once trained and achieve high performance in similar test environments. However, they are often relegated to simulation environments since they require a large amount of training data and can exhibit safety issues during the early training process if trained in the real world. Optimization model-based frameworks are often generalizable to unseen task scenarios (subject to modeling assumptions) and can provide optimality bounds.

Optimization frameworks models the relationship between the tasks and agents.
In \cite{klee2015graph, nicolescu2003natural, galindo2008robot, torreno2017cooperative},
% \cite{klee2015graph, nicolescu2003natural, ekvall2008robot, niekum2012learning, grollman2010incremental, hayes2015effective, galindo2008robot, aeronautiques1998pddl, torreno2017cooperative}
a task is represented as a graph of states (nodes) and actions (edges) so that graph search based algorithms can be applied to find the best sequence of actions for agents.
In \cite{banks2020multi, aksaray2015distributed}, the task constraints are specified using linear temporal logic. Requirements on agent actions are represented as a conjunction of logical expressions.
It is common to assume that agents possess sets of discrete capabilities and that tasks require specific quantities of a given capability to successfully complete an objective. 
The works of \cite{fu2020heterogeneous, fu2021robust} apply such a capability model, where the agent's capabilities are represented as random distributions.
Then, they solve a discrete stochastic optimization to obtain the routes, schedules, and task assignment plans.
The authors of \cite{ravichandar2020strata, prorok2017impact} utilize similar task requirement models, but formulate the problem as an optimal control-style continuous optimization to generate the desired capability distribution at each time step.
A similar capability model is also applied in \cite{messing2022grstaps}.
The authors of \cite{liemhetcharat2011modeling, liemhetcharat2012weighted} represent the dependency between agents in a task using a capability-based graph model and generate a plan using a distributed auction-based method.

Most optimization frameworks for multi-agent task allocation assume the modeling parameters (e.g. task requirements, agent capabilities) are known a priori.
However, this assumption may not hold true in practice where agent capability variations may be hard to predict or determine.
Therefore, the ability to learn the modeling parameters from simulation or experimental data will provide an important contribution to optimization model-based task allocation frameworks.

\subsection{Learning Optimization Models from Data}

Optimization model-based frameworks require a small amount of training data and can be applied to new scenarios with little or no additional training or parameter tuning.
There are many examples from previous works focused on learning the parameters in an optimization model for single-robot manipulation and navigation.
\cite{klee2015graph} and \cite{nicolescu2003natural} leverage a learning from demonstration approach to determine a graph of actions for a new manipulation task. 
Authors of \cite{chou2020explaining, vazquez2018learning} embed linear temporal logic in their optimization model to constrain robot navigation and manipulation trajectories, and develop corresponding methods to learn the temporal logic specifications from demonstrations.

Multi-agent task allocation, on the other hand, has a few previous related works. The authors of \cite{liemhetcharat2011modeling, liemhetcharat2012weighted} model an agent's capabilities as a vector and the dependency between agents for a specific task as a synergy graph. To learn their model, they randomly initialize the synergy graph, and iteratively optimize an objective to estimate the capability values and locally perturb the graph structure to fit the training data.

As mentioned in Sec. \ref{sec:model_based_task_allocation}, the works of \cite{fu2020heterogeneous, fu2021robust, ravichandar2020strata, prorok2017impact, messing2022grstaps} start from the agent capability model and represent the task requirements as a vector of thresholds on each capability type, which are embedded as constraints in the optimization. We extend these models through our proposed framework to enable the learning of agent capability values and task requirement thresholds from simulation or experimental data.
As the learned capability-based model can be regarded as a set of linear constraints, the constraints can then be embedded easily in these existing capability-based models or other optimization-based task allocation frameworks.

\section{Learning Task Requirements and Agent Capabilities as Linear Constraints}\label{sec:learning_model}

The system architecture is shown in Fig. \ref{fig:main_diagram}, suppose the task performance is a function of the team configuration (given a task environment) that can be evaluated through simulation or experiment. In real-world applications, this value can be task completion time, product outputs, or a combined quantity.
We can choose a set of team configurations and evaluate the task performances.
The learning model tries to learn a set of task constraints from these team configurations and performance pairs that can later be embedded in a task allocation model.
This section focuses on learning the task constraints from training data, while the next section provides a task allocation model that utilizes the
learned task constraints.

\subsection{Task Performance Function}
Consider a set of agent types \(K = \{1,\cdots,{\numK{}}\}\), and tasks \(M = \{1, \cdots, {\numM{}}\}\). Suppose \(y_{k i}\) is the number of agents with type \(k \in K\) assigned to a task \(i \in M\). And \(\mathbf{y}_i = [y_{1 i}, \cdots, y_{\numK{} i}]^\transpose \in \mathbb{R}^{\numK{}} \) is the vector for the number of different agents at task \(i \in M\).
Let the task performance be a function of the team configuration: \(f_i(\mathbf{y}_i): \mathbb{R}^{\numK{}} \rightarrow \mathbb{R}\) for each task \(i \in M\).
Ideally, we should set a lower bound, \(f_i^*\), on the performance of a given task as follows
\begin{align}
    f_i(\mathbf{y}_i) \geq f_i^*, \quad \forall i \in M,  \label{eqn:actual_performance_constraint}
\end{align}
and embed it in our task allocation optimization. Depending on the type of performance metric used, the \(\geq\) can be replaced with \(\leq\). However, \(f_i(\mathbf{y}_i)\) is evaluated through an experiment or simulation. It is a non-analytic function and cannot be directly added as a constraint in an optimization. In this section, a set of linear constraints is learned to approximate \eqref{eqn:actual_performance_constraint}.

\subsection{Task Requirements and Agent Capabilities}

\begin{figure}[t]
    \centering
    \includegraphics[width=0.9\linewidth]{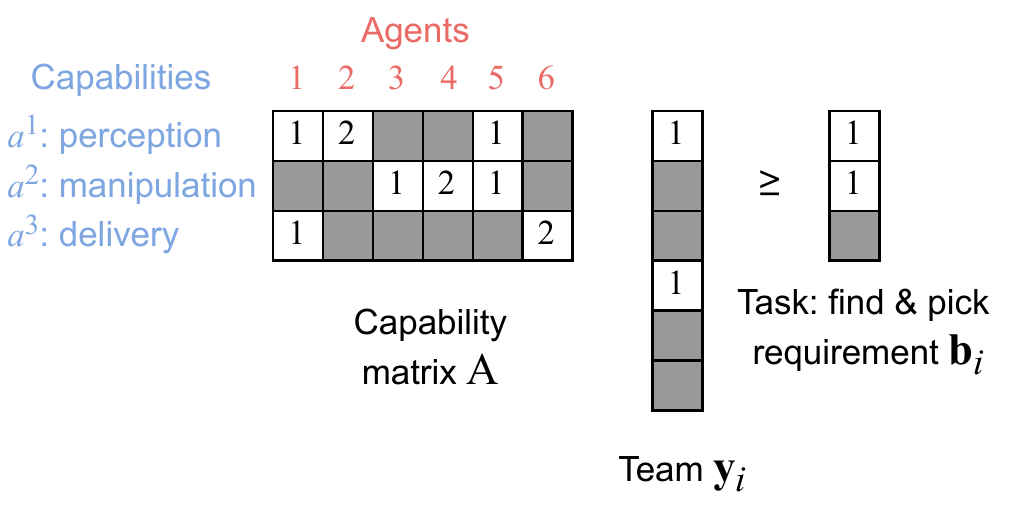}
    \caption{An example team formation problem using a capability matrix and task requirement vectors. Agents can perform three capabilities: perception, manipulation, and delivery. The example task requires perception to explore and locate an object and manipulation to grasp the object.}
    \label{fig:matrix_diagram}
\end{figure}

Consider a set of agent capability types \(C = \{1,\cdots,{\numC{}}\}\) that can be used to characterize an agent's ability to perform a task (e.g., perception, manipulation, or transportation). 
Suppose \(a_{c k}\) represents the value of capability \(c \in C\) that agent type \(k \in K\) possesses. We concatenate these capability values into a capability matrix (see Fig. \ref{fig:matrix_diagram} for example)
\begin{align}
    A &=
    \begin{bmatrix}
    \mathbf{a}_{1} & \cdots & \mathbf{a}_{\numK{}}
    \end{bmatrix} \nonumber \\
   & =
    \begin{bmatrix}
    \mathbf{a}^{1} \\ \vdots \\ \mathbf{a}^{\numC{}}
    \end{bmatrix}
    =
    \begin{bmatrix}
    a_{1 1}   & \cdots & a_{1 \numK{}}  \\
    \vdots    & \ddots & \vdots     \\
    a_{\numC{} 1} & \cdots & a_{\numC{} \numK{}}
    \end{bmatrix}
    \nonumber
\end{align}
where \(\mathbf{a}_{k} = [a_{1 k}, \cdots, a_{\numC{} k}]^\transpose\) is a vector indicating the capabilities of agent type \(k \in K\).

For simplicity, we assume that capabilities are cumulative: if multiple agents work together, the total capability of the agent team is the sum of their individual capability values.
Thus, a team's capability set for a given task can be calculated via matrix multiplication \(A \mathbf{y}_i \in \mathbb{R}^{\numC{}}\). Future work will relax this assumption to allow for non-cumulative capabilities.

Task requirements for a given task \(i \in M\) are provided as a vector of thresholds \(\mathbf{b}_i = [b_{1 i}, \cdots, b_{\numC{} i}]^\transpose\), where \(b_{c i}\) denotes the threshold on capability type \(c \in C\). The model assumes that the task can be successfully completed if the capability requirements are satisfied by the agent team assigned to the task. Mathematically, this requirement can be represented as 
\begin{align}
    A \mathbf{y}_i \geq \mathbf{b}_i, \quad \forall i \in M. \label{eqn:task_constraints}
\end{align}

Equation \eqref{eqn:task_constraints} can be illustrated graphically as in Fig. \ref{fig:matrix_diagram}.

\subsection{Learning the Task Requirements and Agent Capabilities}

With the task requirement model above and the cumulative capability assumption, the actual task constraints in \eqref{eqn:actual_performance_constraint} are approximated using linear constraints in \eqref{eqn:task_constraints}, which can be easily embedded in an optimization program. Ideally, the two constraints should correspond to each other as follows
\begin{align}
    A \mathbf{y}_i \geq \mathbf{b}_i \Leftrightarrow f_i(\mathbf{y}_i) \geq f_i^*, \quad \forall i \in M. \label{eqn:one_to_one_map}
\end{align}

\begin{figure}[t]
    \centering
    \includegraphics[width=0.9\linewidth]{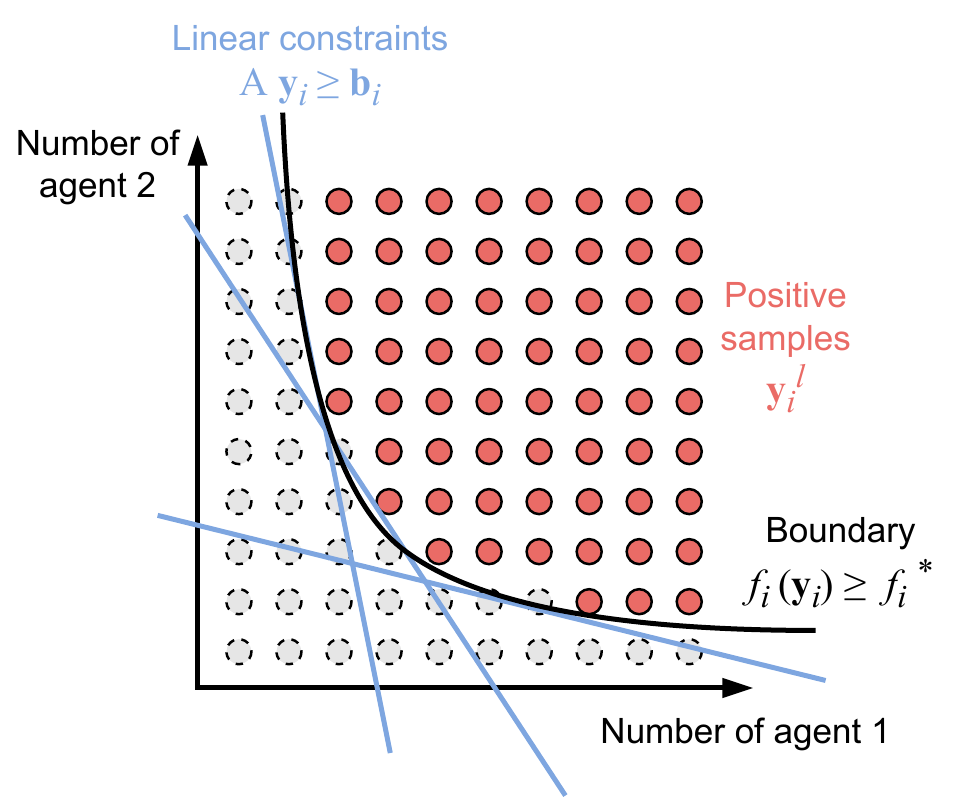}
    \caption{Find the tightest linear boundary for the positive samples.}
    \label{fig:tightest_bound}
\end{figure}

See Fig. \ref{fig:tightest_bound}, since linear constraints can only describe convex polytopes, instead of a one-to-one map in \eqref{eqn:one_to_one_map}, we want to fit the tightest linear boundaries for the region \(f_i(\mathbf{y}_i) \geq f_i^*\).
In practice, \(f_i(\cdot)\) is a non-analytic function evaluated through simulations/experiments, and the region of \(f_i(\mathbf{y}_i) \geq f_i^*\) is not explicitly available. 
Suppose we have evaluated the performance of some team configurations, among which \(\mathbf{y}_i^l, (l = 1, \cdots, n_L)\) satisfy \(f_i(\mathbf{y}_i^l) \geq f_i^*\). We can fit a linear boundary to these positive samples. The following linear program tries to find the \(A\) and \(\mathbf{b}_i\) associated with the tightest linear boundary.
\begin{align}
    \max_{A, \mathbf{b}_i} & \sum_{i \in M} ||\mathbf{b}_i||_1 \label{eqn:basic_learning_model} \\
    \text{sub to } & A \mathbf{y}_i^l \geq \mathbf{b}_i, \ \ \forall l = 1, \cdots, n_L, \ \ \forall i \in M, \nonumber \\
    & ||\mathbf{a}^c||_1 = 1, \ \forall c \in C. \nonumber
\end{align}

In \eqref{eqn:basic_learning_model}, the objective function maximizes the requirement thresholds \(\mathbf{b}_i\) to find the tightest value. Note that \(||\cdot||_1\) stands for the \(L_1\) norm. The first constraint ensures all the positive samples satisfy the constraints defined by \(A\) and \(\mathbf{b}_i\). 
If we scale the capability matrix \(A\) and requirement vector \(\mathbf{b}_i\) by the same value, the constraints \(A \mathbf{y}_i \geq \mathbf{b}_i\) do not change.
As the scale of the capabilities is unconstrained, we add the second constraint in \eqref{eqn:basic_learning_model} to normalize each type of capability.

The optimization problem in \eqref{eqn:basic_learning_model} can further be decomposed row-wise into the following form.
\begin{align}
    & \max_{\mathbf{a}^{c}, b_{c i}} \ \sum_{c \in C} \sum_{i \in M} \mathrm{abs} (b_{c i})  \label{eqn:vector_form_learning_model} \\
    & (\mathbf{a}^c)^\transpose \mathbf{y}_i^l \geq b_{c i}, \ \ \forall l = 1, \cdots, n_L, \ \ \forall i \in M, \ \ \forall c \in C, \nonumber \\
    & ||\mathbf{a}^c||_1 = 1, \ \forall c \in C. \nonumber
    % & \sum_{k \in K} |a_{c k}| = 1, \ \forall c \in C. \nonumber
\end{align}

Note that \(\mathrm{abs}(\cdot)\) denotes the absolute value.
According to \eqref{eqn:vector_form_learning_model}, the optimization of \(\mathbf{a}^{c}\) and \(b_{c i}\) are decoupled for different capabilities \(c \in C\). This is also illustrated in Fig. \ref{fig:matrix_diagram}, as different types of capabilities (rows) are independent of each other. 
% With the assumption that all capability and task threshold are non-negative (all entries in \(A\) and \(\mathbf{b}_i\) are non-negative),
Therefore, we can decompose the optimization in \eqref{eqn:vector_form_learning_model} into \(\numC{}\) separate linear programs as follows

\begin{align}
    & \max_{a_{c k}, b_{c i}} \left(\frac{\alpha_b}{\numM{}} \sum_{i \in M} b_{c i} + \alpha_a \min_{k \in K, (c, k) \in \Acal_1} a_{c k} \right)  \label{eqn:linear_learning_model} \\
    & \sum_{k \in K} {y_{k i}^l \ a_{c k}} \geq b_{c i}, \ \ \forall l = 1, \cdots, n_L, \ \ \forall i \in M, \label{eqn:task_sample_constraint} \\  
    & \sum_{k \in K}{a_{c k}} = 1, \label{eqn:normalization_contraint} \\
    & a_{c k} \geq 0, \quad \forall (c, k) \in \Acal_1, \label{eqn:a1_constraint} \\
    & a_{c k} = 0,    \quad \forall (c, k) \in \Acal_0, \label{eqn:a0_constraint} \\
    & b_{c i} \geq 0, \quad \forall (c, i) \in \Bcal_1, \label{eqn:b1_constraint} \\
    & b_{c i} = 0,    \quad \forall (c, i) \in \Bcal_0. \label{eqn:b0_constraint}
\end{align}

The two constraints in \eqref{eqn:vector_form_learning_model} are rewritten as in \eqref{eqn:task_sample_constraint}-\eqref{eqn:normalization_contraint}. The objective function in \eqref{eqn:vector_form_learning_model} is retained in \eqref{eqn:linear_learning_model} with an additional penalty.
Consider the general case (an example is in Fig. \ref{fig:matrix_diagram}), for which we typically assume that we know an agent's capability types and the task's requirement types, but do not contain knowledge about the specific values\footnote{For example, if a robot contains a manipulator, we know it has a manipulation capability. However, it can be more challenging to quantify the value of that capability in terms of functionality}.
Namely, we know whether an entry in \(A\) and \(\mathbf{b}_i\) is positive or zero (\textbf{the sparsity pattern is known}). Suppose \(\Acal_1 = \{(c, k) | a_{c, k} > 0\}\) and \(\Acal_0 = \{(c, k) | a_{c, k} = 0\}\) are the sets of index pairs for zero and positive agent capabilities, and \(\Bcal_1 = \{(c, i) | b_{c, i} > 0\}\) and \(\Bcal_0 = \{(c, k) | b_{c, i} = 0\}\) are the sets of index pairs for zero and positive task requirements, we express this prior knowledge as constraints \eqref{eqn:a1_constraint}-\eqref{eqn:b0_constraint}.
The right term in \eqref{eqn:linear_learning_model} is added to ensure that positive capabilities are indeed learned to be positive.
The choices of the weights are evaluated through a parametric study, and the results show that the solutions are not sensitive to the weight selections. In general, choosing \(\alpha_a = 0.25 \alpha_b\) would work for most cases with less than 32 capability types.

As a summary, given a set of data evaluated team configurations and the corresponding task performances \((\mathbf{y}_i^l, f_i(\mathbf{y}_i^l)), (l = 1, \cdots, n_L)\), with prior knowledge about whether an entry in the agent capability \(A\) or task requirement \(\mathbf{b}_i\) is non-zero, we can solve \(\numC{}\) small linear programs (with \(\numK{} + \numM{}\) variables) to estimate the linear task requirement constraints for task allocation. As linear programming is polynomially solvable, the learning model developed here is scalable to the number of tasks, agents, and training samples.

\section{Task Allocation with Learned Requirements and Agent Capabilities}\label{sec:task_allocation_model}

Many task allocation frameworks represent the task constraints using a capability-based model \cite{fu2020heterogeneous, fu2021robust, ravichandar2020strata, prorok2017impact}.
In this section, we leverage an optimization model from \cite{fu2021robust} (a simplified version) to show how the learned constraints in Sec. \ref{sec:learning_model} can be embedded in a task allocation framework.

Consider the set of tasks, \(M = \{1, \cdots, {\numM{}}\}\), distributed at different locations, each requiring a team with the correct capabilities to complete the tasks. A set of agent types \(K = \{1,\cdots,{\numK{}}\}\) are located at the depot.
The goal is to complete the tasks using a team of available agents while minimizing the weighted sum of the time and energy spent to travel to and complete the tasks.

\begin{figure}[t!]
	\centering
	\includegraphics[width=0.53\linewidth]{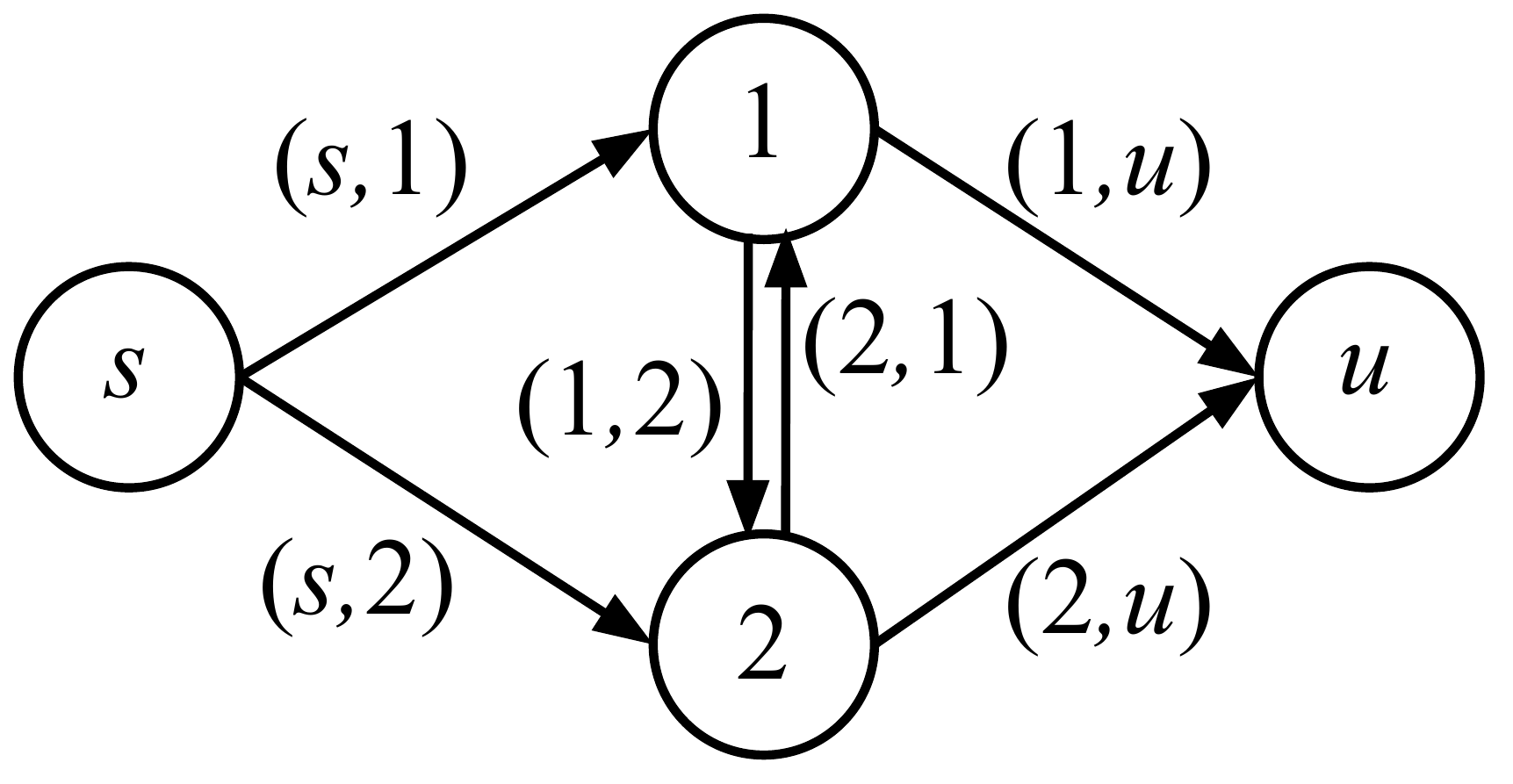}
	\caption{A graphical model example with \(\numM{} = 2\). I.e., there are two task nodes, a start \(s\), and a terminal \(u\).}
	\label{fig:graphical_model}
\end{figure}

\subsection{Graphical Model}

We construct a graph \(G(N,E)\) with the node set \(N = \{1, \cdots, {\numM{}}, s, u\}\) and edge set \(E = \{(i,j) | \forall i,j \in N\}\) (Fig. \ref{fig:graphical_model}). There are edges between every node pair. The node set contains the tasks \(1\) to \(\numM{}\), start \(s = \numM{} + 1\), and terminal nodes \(u = \numM{} + 2\). Note that the integer set \(M\) is a subset of \(N\).
A practical example of this graphical model is shown in the right-most block of Fig. \ref{fig:main_diagram}.
In this graph, the agents depart from the start node, visit all of the tasks assigned to them, and stop at the terminal node.

\subsection{A Mixed-integer Program for Task Allocation}

In this section, we formulate a mixed-integer program for the task allocation problem based on the graphical model in Fig. \ref{fig:graphical_model}. Common notations are explained in Table \ref{tab:variable_definition}. Note that some of the notations have already been defined in Sec. \ref{sec:learning_model}, but are still gathered here for clarity.

\begin{table}[h]
% \normalsize
% \hspace{-0.7cm}
  \caption{Definition of the notations.}
  \label{tab:variable_definition}%
    \begin{tabular}{p{0.06\linewidth}|p{0.8\linewidth}} 
    \toprule
     & Meaning
    \\
    \midrule
% \begin{tabular}{lll} 
    \(x_{k i j}\)& The number of agent \(k \in K\) traveling on edge \((i,j) \in E\).
    \\
    \(y_{k i}\) & The number of agent \(k \in K\) assigned to task \(i \in M\).
    \\
    \(\mathbf{y}_{i}\)& = \([y_{1 i}, \cdots, y_{\numK{} i}]^\transpose \in \mathbb{R}^{\numK{}}\).
    \\
    \(r_{k i j}\)& = 1 if \(x_{k i j} \geq 1\), otherwise 0.
    \\
    \(t_i\)& The time task \(i \in M\) starts or the mission completes (i = u).
    \\
    \(T_{k i j}\)& The time cost for agent \(k\) to travel edge \((i,j)\).
    \\
    \(T_{k i}\)& The time cost for agent \(k\) to complete its part at task \(i \in M\).
    \\
    \(T_{\text{large}}\)& A large time constant.
    \\
    \(d_{k i j}\)& The energy cost for agent \(k\) to travel edge \((i,j)\).
    \\
    \(D_{k}\)& The energy limit for agent \(k\).
    \\
    \(A\) & The learned agent capability matrix.
    \\
    \(\mathbf{b}_i\) & The learned task requirement threshold for task \(i \in M\).
    \\
    \bottomrule
    \end{tabular}
\end{table}

\subsubsection{Objective Function}
In the objective function, we minimize the energy cost and the time to complete all the tasks, where \(C_d\) and \(C_t\) are the weights. The decision variables are \(x\), \(y\), \(r\), and \(t\).
\begin{align}
    \min_{x_{k i j}, y_{k i}, r_{k i j}, t_{i}} C_d \underset{k \in K}{\sum} \ \underset{i,j \in \node}{\sum} d_{k i j} \cdot x_{k i j} + C_t \ t_u
\end{align}

\subsubsection{Task Requirement Constraints} The task requirement constraints are learned according to Sec. \ref{sec:learning_model}.
\begin{align}
    A \mathbf{y}_i \geq \mathbf{b}_i, \quad \forall i \in M. 
    \tag{\ref{eqn:task_constraints}}
\end{align}

\subsubsection{Time Constraints}
The time constraints ensure: first, enough time for the agent to travel the edges; second, all agents arrive before the task starts.
\begin{align}
    t_{i} - t_{j} + T_{k i j} + T_{k i} &\leq T_{\text{large}} (1 - r_{k i j}),
    \forall i , j \in \node, \forall k \in K. \label{eqn:time_constraint1}
\end{align}

\subsubsection{Energy Constraints} The sum energy cost of an agent should be smaller than its limit.
\begin{align}
    \underset{i,j \in \node}{\sum} d_{k i j} \cdot x_{k i j} \leq D_{k}, \quad \forall k \in K.
\end{align}

\subsubsection{Network Flow Constraints}
\eqref{eqn:flow_constraint1} ensures that the agents entering a task also leave after the task is completed.
\eqref{eqn:flow_constraint2} sets an upper bound on the number of agents from each type. \(\numV{}\) is the number of available agents from type \(k\).
\eqref{eqn:node_vehicle_constraint} relates the \(x\) and \(y\) variables: the agent team at a task is comprised of the agents that arrive at the task from all of the edges.
\begin{align}
    \underset{i \in N} {\sum} x_{k i m} &= \underset{j \in N} {\sum} x_{k m j}, \quad \forall m \in M, \ \forall k \in K. \label{eqn:flow_constraint1} \\
    \underset{i \in M} {\sum} x_{k s i} &\leq \numV{}, \quad \quad \quad \forall k \in K. \label{eqn:flow_constraint2} \\
    % y_{k j} &\leq \underset{i \in M} {\sum} x_{k s i}, \quad \forall j \in M, \ \forall k \in K. \label{eqn:flow_constraint3} \\
    y_{k j} &= \underset{i \in N} {\sum} x_{k i j}, \quad \forall j \in M, \ \forall k \in K. \label{eqn:node_vehicle_constraint}
\end{align}

\subsubsection{Helper-variable Constraints}
The following constraint relates the \(x\) and \(r\) variables: \(r_{kij} = 1\) if \(x_{kij} \geq 1\) and \(r_{kij} = 0\) if \(x_{kij} = 0\). \(r_{kij}\) indicates whether there are agents from type \(k\) traveling edge \((i,j)\).
\begin{align}
    \begin{split}
    &x_{kij} \geq r_{kij}, \quad x_{kij} \leq \numV{} \cdot r_{kij}, \quad \forall i,j \in \node, \forall k \in K. \\
    \end{split}\label{eqn:x_r_constraint}
\end{align}    

\subsubsection{Variable Bounds} \(x\), \(y\), and \(t\) are continuous variables, while \(r\) variables are binary.
\begin{align}
    \begin{split}
    &x_{kij} \geq 0, \ y_{ki} \geq 0, \ r_{kij}\in \{0,1\}, \forall i,j \in \node, \forall k \in K. \\
    &t_i \geq 0, \quad \forall i \in N, \quad t_s = 0. \\
    \end{split}\label{eqn:bound_constraint}
\end{align}

Once a solution to the optimization problem is obtained, the variables encode the information about the schedule (\(t\)), routes (\(x\)), and teams (\(y\)) for the tasks.
As an example, in the right-most block in Fig. \ref{fig:main_diagram}, the colored edges and numbers indicate a solution to the \(x\) variables. According to the solution, two large black robots with manipulators first complete the block picking task, and then one of the black robots drives to the exploration task and completes it together with one small white robot.

\section{Computational Evaluation}\label{sec:result_computation}

In this section, we computationally evaluate the scalability and optimality of the algorithm while de-emphasizing the physical meaning of the values involved in the test cases.
All computations are done on a laptop with an Apple M1 chip.
Note that we will show an example of applying our framework to a practical problem in Sec. \ref{sec:result_simulation}.

\subsection{Test Cases}

Assume for some tasks with the performance functions \(f_i(\cdot), i \in M\), the assumption in \eqref{eqn:one_to_one_map} holds. I.e., whether the task can be completed with a team \(\mathbf{y}_i\) can be accurately described as linear constraints.
We use the following \(f_i(\mathbf{y}_i)\) to replace the simulation or experiment that evaluates the performance of a team \(\mathbf{y}_i\) during the computation evaluation.
\begin{align}
    f_i(\mathbf{y}_i) =
    \begin{cases} 1, \quad A^g \mathbf{y}_i \geq \mathbf{b}_i^g \\
    0, \quad \textnormal{otherwise}
    \end{cases}
    \forall i \in M. \label{eqn:computational_function}
\end{align}

\(A^g\) and \(\mathbf{b}_i^g\) are the preset ground truth capabilities and task requirements and are unknown to the learning model. The learning model should estimate \(A\) and \(\mathbf{b}_i\) which describe the same constraints as \eqref{eqn:computational_function}.

We define eight test cases of different sizes (given in \tableref{} \ref{tab:random_test_cases}) and test the prediction accuracy and computational time of the learning model.
For each row in \tableref{} \ref{tab:random_test_cases}, we randomly initialize the ground truth matrices \(A^g\) and \(\mathbf{b}_i^g\) in \eqref{eqn:computational_function} based on the selected hyper-parameters and generate the data \((\mathbf{y}_i^l, f_i(\mathbf{y}_i^l)), l = 1, \cdots, n_L\).
For a particular task, the team configurations exist within \(\mathbf{y}_i \in [0, n_s]^{\numK{}}\),  where \(n_s\) is the number of agents within each type. As an example, if \(n_s = 5\) and there are \(\numK{} = 6\) related agent types, the size of the configuration space is \(46656 = (5+1)^6\).
% Note that \((n_s+1)^{\numK{}}\) is kept small such that all possible team configurations can be evaluated.

% Table generated by Excel2LaTeX from sheet 'Sheet3'
\begin{table}[t]
  \centering
  \caption{Test cases of different sizes and the training time. \(\numM{}\) is the task number. \(\numK{}\) is the number of agent types. \(n_s\) is the number of agents within each type. \(\numC{}\) is the number of capabilities. \(\bar{n}_L\) is the mean size of the team configuration space for the tasks, while \(\sum n_L\) is the sum. Entire: the average training time using the entire team configuration space. Random: the average training time using a random subset of team configurations.}
    \begin{tabular}{ccccccc|cc}
    \toprule
    Case  & \(\numM{}\) & \(\numK{}\) & \(\numC{}\) & \(n_s\) & \(\bar{n}_L\) & \(\sum n_L\) & Entire & Random \\
    \midrule
    0     & 8     & 6     & 8     & 5     & 1.5k  & 12k   & 5.8   & 0.6 \\
    1     & 8     & 6     & 8     & 5     & 6k    & 48k   & 20.7  & 0.5 \\
    2     & 8     & 6     & 16    & 5     & 6k    & 48k   & 51.2  & 1.1 \\
    3     & 8     & 6     & 32    & 5     & 6k    & 48k   & 104.7 & 2.6 \\
    4     & 20    & 6     & 8     & 5     & 7.5k  & 150k  & 59.8  & 1.4 \\
    5     & 40    & 6     & 8     & 5     & 7.5k  & 300k  & 108.4 & 2.7 \\
    6     & 40    & 6     & 16    & 5     & 7.5k  & 300k  & 275.1 & 5.9 \\
    7     & 40    & 6     & 32    & 5     & 7.5k  & 300k  & 542.1 & 12.0 \\
    \bottomrule
    \end{tabular}%
  \label{tab:random_test_cases}%
\end{table}%

\subsection{Accuracy and Computational Cost}

For each test case size in \tableref{} \ref{tab:random_test_cases}, we randomly generate 10 realizations and applied the learning model to estimate the capabilities and task requirements.
The learned values of \(A\) and \(\mathbf{b}_i\) are then used to predict whether a team can complete a task according to \eqref{eqn:computational_function}. The prediction is compared with the label generated by the ground truth \(A^g\) and \(\mathbf{b}_i^g\).

For a real-world problem, obtaining team performance samples can be expensive (through experiments/simulations), and evaluating the entire team configuration space for estimating the capability and requirements can be impractical. To investigate whether the model learns with a small training set, we evaluate the algorithm in two modes: 1) training and testing both on the entire team configuration space (up to 46656 samples for a task); 2) training on at most 200 randomly selected team configurations for each task while testing across the entire configuration space.

\begin{figure}[t]
    \centering
    \includegraphics[width=0.8\linewidth]{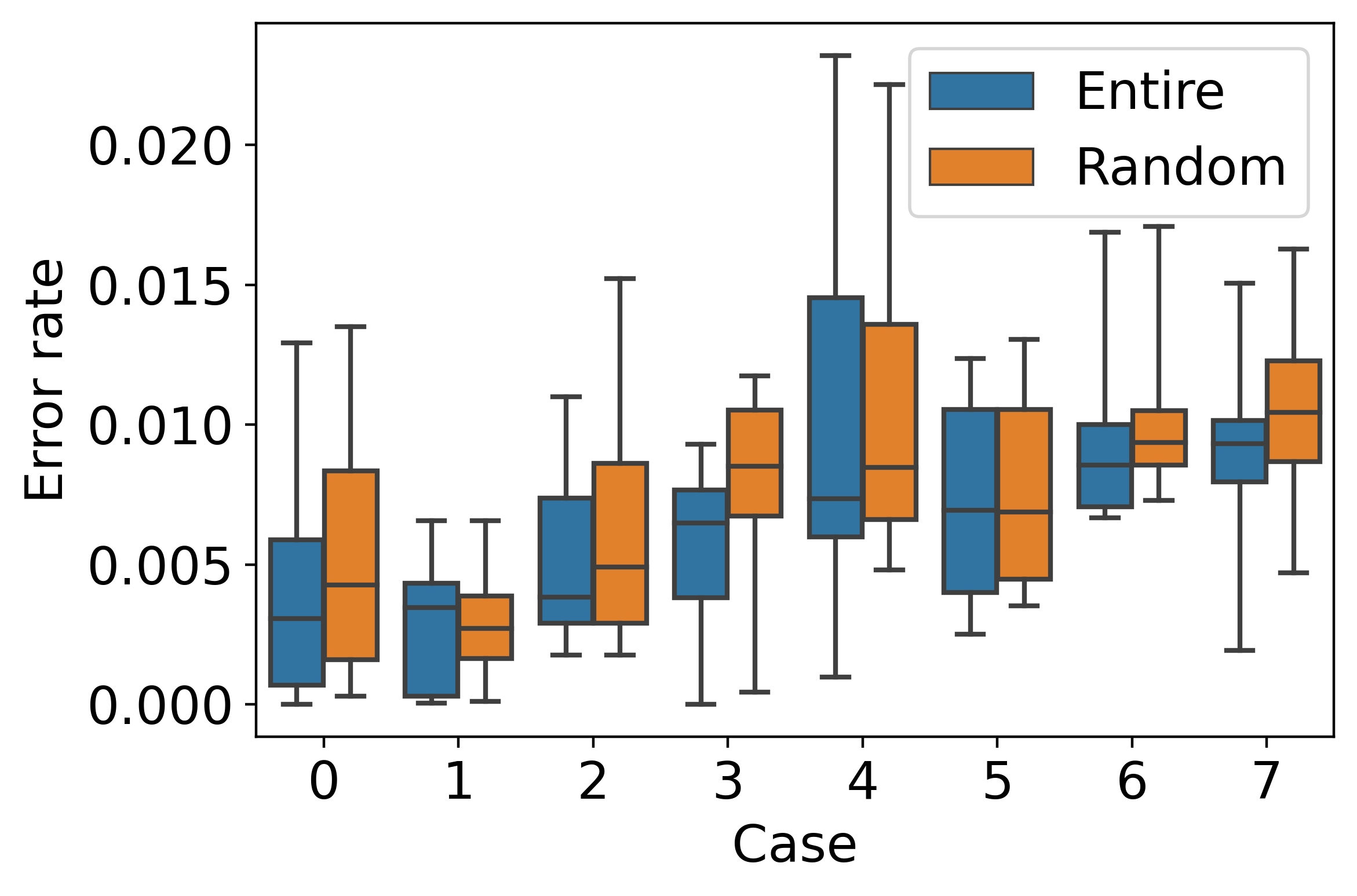}
    \caption{Prediction error tested on the complete data. `Entire' refers to the result trained with the entire configuration space (up to 46656 teams), while random contains a subset of the configuration space.}
    \label{fig:computational_error}
\end{figure}

The distribution of prediction errors are shown in Fig. \ref{fig:computational_error}.
The prediction error is around 0.5-2\%. This shows that our algorithm can accurately recover the capability if the modeling assumptions are satisfied.
The test errors of the results trained on 200 randomly selected samples are only slightly larger than the model trained on the entire data set. 
Therefore, the task requirements and agent capabilities can be estimated using a small percentage of the data within the configuration space.

The learning model assumes the sparsity pattern is known.
However, we also evaluate the situations when we have inaccurate sparsity knowledge. The prediction errors do increase, but capabilities can still be estimated. For example, a 10\% error in the sparsity matrix results in 2-6\% prediction errors for the eight cases.

The average training times are listed in \tableref{} \ref{tab:random_test_cases}.
The computational time grows linearly with respect to the number of tasks and capabilities. For case 7 (the largest case), the average training times using the entire configuration space versus randomly selected subsets are 542 and 12 seconds, respectively. From this analysis, the learning model can easily scale to a mission comprised of 40 subtasks and 32 different capabilities. 
Considering the training time needed, the model can update the capability and requirement values every few seconds for real-time applications.

\section{Simulation Application}\label{sec:result_simulation}

\begin{figure}[t]
    \centering
    \includegraphics[width=0.8\linewidth]{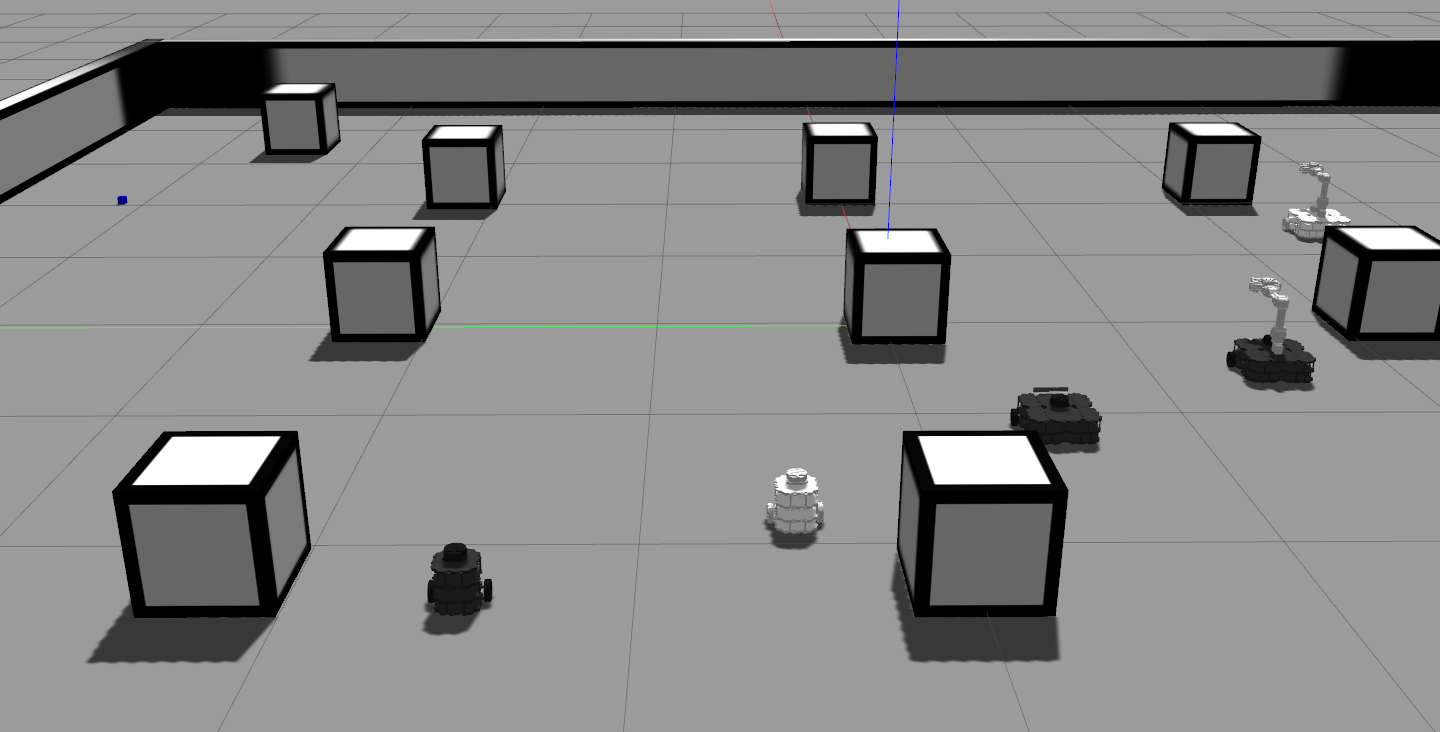}
    \caption{A screenshot of the multi-agent simulation environment. The robot team is conducting a `find and pick' task.
    }
    \label{fig:simulation_environment}
\end{figure}

In this section, we apply our learning framework to a robotic exploration and manipulation problem and show how the agent capabilities and task requirements are learned. 
Multiple types of tasks are involved in the example to show that the framework generalizes to different tasks.

The simulation environment is developed in Gazebo, and the communication between robots and sensors is established using ROS. A screenshot is shown in Fig. \ref{fig:simulation_environment}. There are five types of agents, and their qualitative capabilities are described in \tableref{} \ref{tab:real_agent_capability}. There are five types of tasks as shown in \tableref{} \ref{tab:real_task_requirement}.
Once the team configuration is determined for a specific task, a simple intra-team strategy (greedy algorithms) will guide the agents to complete the task cooperatively. As this paper focuses on inter-team scheduling and team configuration planning, the simple intra-team strategy is not described here.

\begin{table}[t]
  \centering
  \caption{Qualitative agent capabilities. A screenshot of the five robots is shown in Fig. \ref{fig:simulation_environment} (left to right: agent type 1-5). }
    \begin{tabular}{l|p{0.25\linewidth}p{0.45\linewidth}}
    \toprule
    \multicolumn{1}{c|}{Agents} & Perception & Manipulation \\
    \midrule
    1: Smallbot 1 & The range is 1 m.     & None. \\
    2: Smallbot 2 & The range is 1.5 m.     & None. \\
    3: Largebot 1 & The range is 1 m.     & None. \\
    4: Largebot 2 & The range is 1 m.     & Only picks light blocks. Picks light blocks faster than largebot 3. \\
    5: Largebot 3 & The range is 1.5 m.     & Picks both light and heavy blocks. \\
    \bottomrule
    \end{tabular}%
  \label{tab:real_agent_capability}%
\end{table}%

\begin{table}[t]
  \centering
  \caption{Qualitative task requirements. Perception is not required for the picking tasks as the block locations are known. The size of a task region is set to 6\(\times\)6 meters.}
    \begin{tabular}{l|p{0.25\linewidth}p{0.4\linewidth}}
    \toprule
    \multicolumn{1}{c|}{Tasks} & Perception & Manipulation \\
    \midrule
    1: Explore & Cover the region.     & None. \\
    2: Pick (light) & None.     & Pick four light blocks at known locations. \\
    3: Pick (mixed) & None.     & Pick two light and two heavy blocks at known locations. \\
    4: Pick (heavy) & None.     & Pick four heavy blocks at known locations. \\
    5: Find and pick & Find a block.     & Pick the found light block. \\
    \bottomrule
    \end{tabular}%
  \label{tab:real_task_requirement}%
\end{table}%

\subsection{Learning the Capability Model}

Considering the physical size of the task area, we add a constraint that a team for a task can contain at most four agents. There are 125 valid teams within the configuration space of \(\mathbb{N}^5\).
Considering the additional requirements\footnote{Task 2 is only relevant to the largebots 2 and 3. Task 3 requires at least one largebot 3 in the team. Task 4 is only relevant to largebot 3. Task 5 requires at least one agent with the manipulation capability.}, the number of valid teams for the five tasks are 125, 14, 10, 4, and 91, respectively. Among these possible team configurations, we chose to evaluate the performance of 20, 14, 10, 4, and 25 team configurations within the five tasks.

For each task \(i = 1, \cdots, 5\) and team configuration \(\mathbf{y}_i^l\) (\(l\) is the label for a team configuration), define the task performance as \(f_i(\mathbf{y}_i)\). Here, it is the task completion time.
\(f_i(\mathbf{y}_i)\) is stochastic because the completion time is influenced by the localization, control, communication uncertainties, and environmental setup.
Therefore, instead of \(f_i(\mathbf{y}_i) \leq f_i^*\), we use \eqref{eqn:prob_task_requirement} to determine whether a team is valid.
\begin{align}
    P(f_i(\mathbf{y}_i) \leq f_i^*) \geq 80 \% \label{eqn:prob_task_requirement}
\end{align}
To check the condition in \eqref{eqn:prob_task_requirement}, we randomly generate five cases for each \((i, l)\) pair (with different block locations) and evaluate the task completion time. If at least four of the task completion times are smaller than the task time threshold \(f_i^*\), the team is considered valid.
The statistics of the task completion time are listed in \tableref{} \ref{tab:task_statistics}.

% Table generated by Excel2LaTeX from sheet 'paper'
\begin{table}[t]
  \centering
  \caption{Task completion time and threshold selection for the tasks. The unit for the \{min, max, mean, thres\} are seconds. A specific team configuration \(\mathbf{y}_i\) is considered valid if the completion time \(f_i(\mathbf{y}_i)\) is smaller than the threshold. The error rate calculates the proportion of falsely predicted samples using the learning model.}
    \begin{tabular}{l|ccccc|c}
    \toprule
    \multicolumn{1}{c|}{Tasks} & Sample & Min   & Max   & Mean  & Thres & Error \\
    \midrule
    1: Explore & 20    & 85    & 245   & 156   & 190   & 0\% \\
    2: Pick (light) & 14    & 44    & 211   & 99    & 110   & 0\% \\
    3: Pick (mixed) & 10    & 66    & 252   & 117   & 130   & 0\% \\
    4: Pick (heavy) & 4     & 70    & 203   & 140   & 140   & 0\% \\
    5: Find and pick & 25    & 44    & 111   & 74    & 90    & 4\% \\
    \bottomrule
    \end{tabular}%
  \label{tab:task_statistics}%
\end{table}%

We apply the learning model described in Sec. \ref{sec:learning_model} to this case study to learn a set of agent capabilities and task requirements to approximate the boundary between valid and invalid teams.
We define five capability types (1: perception, 2: manipulation related to light blocks,  3: manipulation related to heavy blocks, 4: perception 2, 5: manipulation 2 related to light blocks).
The prior information is the sparsity pattern of the capability and requirement matrices:
which values in \tableref{}s \ref{tab:learned_cap}-\ref{tab:learned_req} are nonzero.
Note that capabilities 4-5 are copies of 1-2 related to the find-and-pick task. Adding the two additional capabilities makes the capability-based model more expressive and results in lower prediction error.
The practical meaning is that a specific capability of an agent (e.g. perception) can change with regard to different tasks.

The learned values with the threshold listed in \tableref{} \ref{tab:task_statistics} are shown in \tableref{}s \ref{tab:learned_cap}-\ref{tab:learned_req}. The values indicate that 1) smallbot2 and largebot3 have larger perception capabilities and 2) largebot 2 has a larger manipulation capability when picking light blocks. This is consistent with their qualitative capability in \tableref{} \ref{tab:real_agent_capability}.
The prediction error with the selected threshold is shown in \tableref{} \ref{tab:task_statistics}. A parametric study has been done and shown that similar capability and task requirements values can be learned with an over \(\pm 50\) perturbation of the selected threshold.

% Table generated by Excel2LaTeX from sheet 'paper'
\begin{table}[t]
  \centering
  \caption{Learned agent capabilities \(A^\transpose\).}
    \begin{tabular}{c|ccccc}
    \toprule
    Agents & 1: P  & 2: ML & 3: MH & 4: P2 & 5: ML2 \\
    \midrule
    1: Smallbot 1 & 1     & \cellcolor[rgb]{ .647,  .647,  .647}0 & \cellcolor[rgb]{ .647,  .647,  .647}0 & 1     & \cellcolor[rgb]{ .647,  .647,  .647}0 \\
    2: Smallbot 2 & 2     & \cellcolor[rgb]{ .647,  .647,  .647}0 & \cellcolor[rgb]{ .647,  .647,  .647}0 & 1     & \cellcolor[rgb]{ .647,  .647,  .647}0 \\
    3: Largebot 1 & 1     & \cellcolor[rgb]{ .647,  .647,  .647}0 & \cellcolor[rgb]{ .647,  .647,  .647}0 & 1     & \cellcolor[rgb]{ .647,  .647,  .647}0 \\
    4: Largebot 2 & 1     & 2     & \cellcolor[rgb]{ .647,  .647,  .647}0 & 1     & 1 \\
    5: Largebot 3 & 2     & 1     & 1     & 1     & 1 \\
    \bottomrule
    \end{tabular}%
    \vspace{-0.2cm}
  \label{tab:learned_cap}%
\end{table}%

% Table generated by Excel2LaTeX from sheet 'paper'
\begin{table}[t]
  \centering
  \caption{Learned task requirements. Each row indicates a vector \(\mathbf{b}_i^\transpose\).}
    \begin{tabular}{l|ccccc}
    \toprule
    \multicolumn{1}{c|}{Tasks} & 1: P  & 2: ML & 3: MH & 4: P2 & 5: ML2 \\
    \midrule
    1: Explore & 2     & \cellcolor[rgb]{ .647,  .647,  .647}0 & \cellcolor[rgb]{ .647,  .647,  .647}0 & \cellcolor[rgb]{ .647,  .647,  .647}0 & \cellcolor[rgb]{ .647,  .647,  .647}0 \\
    2: Pick (light) & \cellcolor[rgb]{ .647,  .647,  .647}0 & 4     & \cellcolor[rgb]{ .647,  .647,  .647}0 & \cellcolor[rgb]{ .647,  .647,  .647}0 & \cellcolor[rgb]{ .647,  .647,  .647}0 \\
    3: Pick (mixed) & \cellcolor[rgb]{ .647,  .647,  .647}0 & 3     & 1     & \cellcolor[rgb]{ .647,  .647,  .647}0 & \cellcolor[rgb]{ .647,  .647,  .647}0 \\
    4: Pick (heavy) & \cellcolor[rgb]{ .647,  .647,  .647}0 & \cellcolor[rgb]{ .647,  .647,  .647}0 & 3     & \cellcolor[rgb]{ .647,  .647,  .647}0 & \cellcolor[rgb]{ .647,  .647,  .647}0 \\
    5: Find and pick & \cellcolor[rgb]{ .647,  .647,  .647}0 & \cellcolor[rgb]{ .647,  .647,  .647}0 & \cellcolor[rgb]{ .647,  .647,  .647}0 & 2     & 1 \\
    \bottomrule
    \end{tabular}%
  \label{tab:learned_req}%
\end{table}%

In summary, the learning model can be applied to a practical problem to learn meaningful capabilities and task requirements with low prediction errors. The learning process is not sensitive to threshold selection.

\subsection{Task Allocation with the Learned Capabilities}\label{sec:result_task_allocation_simulation}

In this section, we insert the learned agent capabilities \(A\) and task requirements \(\mathbf{b}_i\) into the task allocation framework described in Sec. \ref{sec:task_allocation_model} and solve a practical allocation problem to generate the team, routes, and schedules for the tasks.

\begin{figure}[t]
    \centering
    \includegraphics[width=0.9\linewidth]{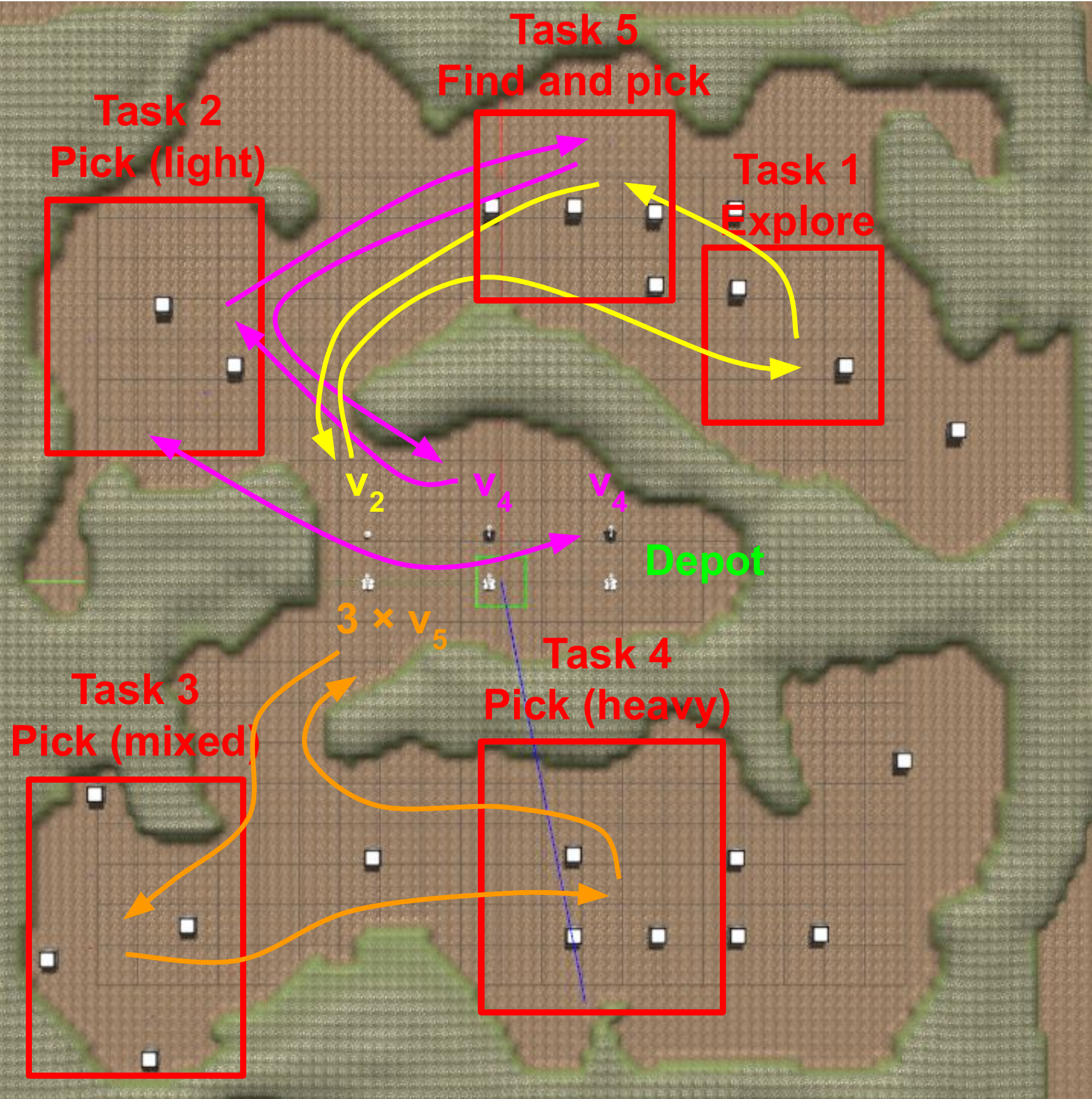}
    \caption{A multi-agent task allocation problem consisting of five heterogeneous sub-tasks. An agent from type \(i\) is marked as \(v_i\) in the figure. The teams for the tasks are shown as follows. Task 1: \(v_2\), task 2: two \(v_4\), task 3 and 4: three \(v_5\), task 5: one \(v_2\) and one \(v_4\).
    }
    \label{fig:task_allocation_simulation}
\end{figure}

In Fig. \ref{fig:task_allocation_simulation}, five heterogeneous tasks (chosen from \tableref{} \ref{tab:real_task_requirement}) are distributed in a 30\(\times\)30 meters outdoor area. The tasks should be completed within a required time duration once proper agent teams reach the task region. The shortest paths to travel between any two task locations are computed and the corresponding energy and time cost are generated. We have four agents from all types in \tableref{} \ref{tab:real_agent_capability} and the team size for any task should be smaller than four. 

By applying the task allocation model in Sec. \ref{sec:task_allocation_model} using the learned agent capabilities and requirements in \tableref{}s \ref{tab:learned_cap}-\ref{tab:learned_req}, a teaming plan is generated to minimize a weighted sum of energy and time while satisfying all task requirements. The planned routes and teams are shown in Fig. \ref{fig:task_allocation_simulation}. According to the solution, the planner chooses one agent 2, two agents 4, and three agents 5 to complete the five tasks.

The teaming and routing plan in Fig. \ref{fig:task_allocation_simulation} is evaluated in simulation.
The total time for % the
agents to complete all tasks and travel back to the depot is 317 seconds.
The task time threshold in \tableref{} \ref{tab:task_statistics} are all maintained.
As the task constraints are learned indoor (Fig. \ref{fig:simulation_environment}) and applied in a slightly different setup in an outdoor region, it shows the learned constraints can be applied to similar but different setups. Overall, the results validate that the task requirement learning and task allocation framework can work together and be applied to solve practical problems in simulation.

\section{Conclusions and Future Work}\label{sec:conclusion}

\subsection{Conclusions}

Many optimization model-based multi-agent task allocation frameworks implement an agent capability and task requirement model, where the capability and requirement parameters are assumed given.
This paper addresses the estimation problem of the capability and requirement parameters in such models and, therefore, facilitates the application of these model-based frameworks to practical problems.

The parameter estimation problem is formulated as a linear program that tries to find the agent capabilities and task requirements that optimally fit the team configuration and task performance pairs.
For situations where the team configuration space is huge and an exhaustive performance evaluation is impractical, a randomly chosen subset provides sufficient information for the estimation.
A comprehensive computational evaluation shows that the learning model can fit the data with low errors (\(\leq\) 2\% for most cases) and short training times (a few seconds with randomly chosen data).

A Gazebo-based simulation platform is developed, and the learning framework is validated in a practical multi-agent problem.
Using the task completion time obtained from the simulation as the performance metric, the agent and task setups in \tableref{}s \ref{tab:real_agent_capability}-\ref{tab:real_task_requirement} can be successfully reflected by the learned values.
Finally, the learned capabilities and task requirements are embedded as constraints in a task allocation optimization, and the system is validated in a practical multi-agent exploration and manipulation example (Sec. \ref{sec:result_task_allocation_simulation}).

\subsection{Limitations and Future Work}

First, the capability model in this paper assumes all agent capabilities are cumulative which results in a linear representation. In practice, there can be non-cumulative capabilities. For example, the speed that a team can drive is not the sum of the agent speeds. Additional modeling can be performed to learn and apply non-cumulative capabilities for task allocation.
Second, the task performance of a team configuration can vary due to the disturbances and uncertainties in the system.
Future work will consider developing learning models that encode and estimate such uncertainties in the agents' task capabilities.
In addition, experimental verification will be conducted in the real world for the framework developed in this paper.

\bibliography{style/ieee-abrv,style/strings-abrv,reference/ref,ms}
\bibliographystyle{IEEEtran}

\end{document}